\documentclass{vgtc}                          % final (conference style)

\ifpdf%                                % if we use pdflatex
  \pdfoutput=1\relax                   % create PDFs from pdfLaTeX
  \usepackage{graphicx}                % allow us to embed graphics files
  \DeclareGraphicsExtensions{.pdf,.png,.jpg,.jpeg} % for pdflatex we expect .pdf, .png, or .jpg files
\else%                                 % else we use pure latex
  \usepackage{graphicx}                % allow us to embed graphics files
  \DeclareGraphicsExtensions{.eps}     % for pure latex we expect eps files
\fi%

\graphicspath{{figures/}{pictures/}{images/}{./}} % where to search for the images

\usepackage{microtype}                 % use micro-typography (slightly more compact, better to read)
\PassOptionsToPackage{warn}{textcomp}  % to address font issues with \textrightarrow
\usepackage{textcomp}                  % use better special symbols
\usepackage{mathptmx}                  % use matching math font
\usepackage{times}                     % we use Times as the main font
\usepackage{cite}                      % needed to automatically sort the references
\usepackage{tabu}                      % only used for the table example
\usepackage{booktabs}                  % only used for the table example
\usepackage{amsmath}
\usepackage{graphicx}	
\usepackage{amsmath}	
\usepackage{amssymb}	
\usepackage{booktabs}
\usepackage{times}
\usepackage{microtype}
\usepackage{epsfig}
\usepackage[table,xcdraw,dvipsnames]{xcolor}
\usepackage{caption}
\usepackage{float}
\usepackage{placeins}
\usepackage{color, colortbl}
\usepackage{stfloats}
\usepackage{enumitem}
\usepackage{tabularx}
\usepackage{xstring}
\usepackage{multirow}
\usepackage{xspace}
\usepackage{url}
\usepackage{subcaption}
\usepackage{xcolor}
\usepackage[hang,flushmargin]{footmisc}
\usepackage{footnote}
\usepackage{graphicx}
\usepackage{amsmath}
\usepackage{amssymb}
\usepackage{booktabs}
\usepackage{array}
\usepackage{multirow}
\usepackage{makecell}
\usepackage{bbding}
\PassOptionsToPackage{normalem}{ulem}
\usepackage{ulem}
\usepackage{arydshln}
\usepackage{xcolor}         % colors
\usepackage[accsupp]{axessibility}
\usepackage{pifont}
\usepackage{combelow}  %%miya
\usepackage{hyperref}
\usepackage{graphicx}
\usepackage{amssymb}
\usepackage{xcolor}
\usepackage{pifont}
\usepackage{multirow}

\usepackage[numbers]{natbib}
\newcommand{\eg}{e.\,g.\ }

\onlineid{0}

\vgtccategory{Research}

\title{LiDAR-Forest Dataset: LiDAR Point Cloud Simulation Dataset for \\ Forestry Application}

\newcommand{\authors}{
    {\textsuperscript{1}} Yawen \ Lu\ \thanks{e-mail: lu976@purdue.edu},\ \
    {\textsuperscript{1}} Zhuoyang \ Sun\ \thanks{e-mail: sun1233@purdue.edu}, \ \
    {\textsuperscript{2}} Jinyuan \ Shao\ \thanks{e-mail: jyshao@purdue.edu}, \ \
    {\textsuperscript{1}} Qianyu \ Guo\ \thanks{e-mail: guo716@purdue.edu}, \ \
    {\textsuperscript{1}} Yunhan \ Huang\ \thanks{e-mail: huan1482@purdue.edu}, \ \
    {\textsuperscript{2}} Songlin \ Fei\thanks{e-mail: sfei@purdue.edu}, \ \
    {\textsuperscript{1}} Victor \ Chen\ \thanks{e-mail: victorchen@purdue.edu} \ \
}

\author{\authors\\
     \scriptsize \textsuperscript{1}Polytechnic Institute, Purdue University, USA \ \
     \scriptsize \textsuperscript{2}Department of Forestry and Natural Resources, Purdue University, USA 
}

\begin{document}

%% Abstract section.
\abstract{The popularity of LiDAR devices and sensor technology has gradually empowered users from autonomous driving to forest monitoring, and research on 3D LiDAR has made remarkable progress over the years. Unlike 2D images, whose focused area is visible and rich in texture information, understanding the point distribution can help companies and researchers find better ways to develop point-based 3D applications. In this work, we contribute an unreal-based LiDAR simulation tool and a 3D simulation dataset named \textbf{\textit{LiDAR-Forest}}, which can be used by various studies to evaluate forest reconstruction, tree DBH estimation, and point cloud compression for easy visualization. The simulation is customizable in tree species, LiDAR types and scene generation, with low cost and high efficiency.}

\vspace{-1mm}
%%% Refer to https://dl.acm.org/ccs
\CCScatlist{
  \CCScatTwelve{Computing methodologies}{Modeling and simulation}{Simulation support systems}{Simulation environments};
  \CCScatTwelve{Computing methodologies}{Computer graphics}{Shape modeling}{Point-based models}.
}

%%%%%%%%%%%%%%%%%%%%%%%%%%%%%%%%%%%%%%%%%%%%%%%%%%%%%%%%%%%%%%%%
%%%%%%%%%%%%%%%%%%%%%% START OF THE PAPER %%%%%%%%%%%%%%%%%%%%%%
%%%%%%%%%%%%%%%%%%%%%%%%%%%%%%%%%%%%%%%%%%%%%%%%%%%%%%%%%%%%%%%%%

%% \begin{document}

%% The ``\maketitle'' command must be the first command after the
%% ``\begin{document}'' command. It prepares and prints the title block.

%% the only exception to this rule is the \firstsection command
%\firstsection{Introduction}

\maketitle

\section{Introduction} %for journal use above \firstsection{..} instead

On a cold winter night, you carry a backpack with a 64-bit LiDAR sensor to collect forest data such as LiDAR points and IMU. Because the wild forest is so dense and large, with extremely rugged ground full of shrubs, it is easy to get scratched or lost and not know where you are or where you have already recorded. But this is just a small area, and your goal is to cover as much diversity, weather, and area as possible, which could take half a year. You complain about whatever and go back to sleep, hoping everything will be fine tomorrow.

\begin{figure}[htb]
\centering
\includegraphics[width=1.0\linewidth, height=3.3cm]{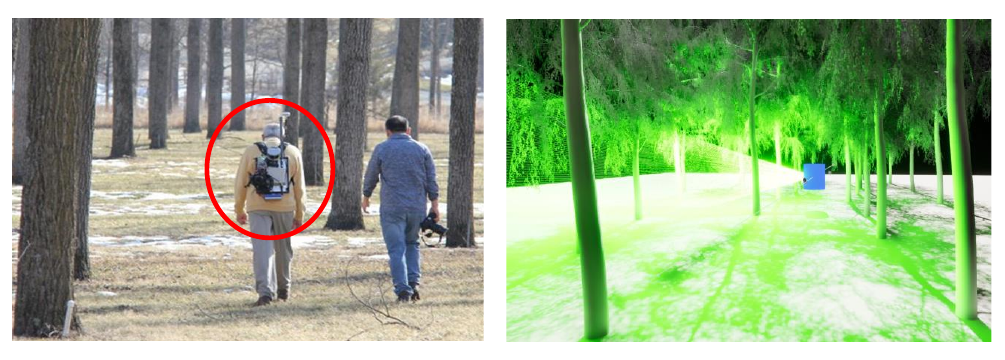}
\vspace{-5mm}
\caption{Conceptual difference between the real-word LiDAR backpack scanner (left) and our simulated LiDAR scanner in wild forests (right).
\label{fig:intro}}
\vspace{-0.1in}
\end{figure}

It is a miniature of the forestry researchers and professionals who want to collect point cloud in wild forests. When comparing with the most recent equipment for collecting LiDAR point in forest scenarios (see Figure~\ref{fig:intro}), we naturally raise three questions: \textbf{i)} Is it possible to use simulation data as an efficient complement when evaluating point cloud algorithms?
\textbf{ii)} How can we guarantee that the simulation is close to the data collected by the real LiDAR sensor?
\textbf{iii)} What applications can this simulation dataset be used for that will benefit forestry professionals and education?

Real-world collection in a structured environment, such as a track of planted forests, allows end-to-end testing of the developed system, but it is limited to a very small number of test cases because it is very expensive and time-consuming to cover diverse tree species and regions. In addition, manual measurement and data annotation relies on post-processing by human labelers, which inevitably introduces subjective errors and noise into the training labels, especially for trees that are out of range of the LiDAR (either too close or too far away). Given these limitations, a simulation system for LiDAR scanning is necessary and important, to automatically generate high-quality, accurate data with error-free labels.

There have already been some attempts to recognize the importance of sensor simulation, dating back to the early robot simulators, such as Gazebo and OpenRave \cite{diankov2008openrave, koenig2004design}, which support sensor simulation through physics and graphics engines. More recently, advanced real-time rendering techniques have been used in autonomous driving simulators such as CARLA and AirSim \cite{dosovitskiy2017carla, shah2018airsim}. However, these existing simulators primarily focus on urban driving scenarios, relying on manually created 3D assets and simplified physics, resulting in simulations that look fake and are very limited in variety. This limits the wide use of existing methods to be applied in wild forests.

Bridging the gap between simulation and the real world requires us to design a better model of the real world environment and the distribution of the LiDAR point cloud. In this work, to address the three questions we raised earlier, we propose \textbf{\textit{LiDAR-Forest}} dataset, which focuses on the LiDAR point cloud simulation for the challenging wild forest scenes. As a new, efficient, and realistic simulation system, it simulates the actual LiDAR in a more realistic manner. The entire system consists of \textit{three stages and five novel modules}: The asset creation stage creates each core component such as LiDAR, trees of different species, and a customizable landscape. The scene generation stage combines the individual components together with a random selector for the number and location distribution. At the simulation stage, our approach combines the power of physics-based scenes and LiDAR sensor and records the generated point cloud to the drive. The five modules (data generation module, rotation module, error simulation module, human motion module, and scan path module) contribute to a more realistic simulation by introducing motion and sensor noises, and narrow the gap between the simulation and the real-world backpack collection.

In the following section, we will describe the background of LiDAR simulation and the relevant LiDAR-based applications in forestry in Sec.~\ref{sec:background}, the design and creation of our dataset and metrics in Sec.~\ref{sec:dataset}, the extensibility and potential applications in Sec.~\ref{sec:usage}, a discussion of future work in Sec.~\ref{sec:future}, and a conclusion summarizing the work in Sec.~\ref{sec:conclusion}. For its effectiveness, we hope the simulation system and data can catalyze a transformation in simulation systems and inspire new insights to the digital forestry community.

\section{Background}
\label{sec:background}

%%%%%%%%%%%%%%%%%
%  From Jinyuan %
LiDAR point clouds have been widely used for various tasks in forestry including tree counting, species identification, DBH measurements, volume estimation, and structural diversity quantification.
In this section, we generally review recently released
LiDAR point cloud datasets in forestry, followed by introducing the recent simulation systems. Finally, we summarize the LiDAR-based applications.

\textbf{LiDAR Dataset for Forestry.} A LiDAR sensor is capable of measuring the distance data of a set of adjacent points using laser beams and generating a point cloud in a 3D manner. Considering it's high measurement accuracy, data collected by LiDAR sensor has been widely used in the research of forestry. \citet{weiss2011plant} proposed a LiDAR-based plant detection and segmentation approach, while demonstrating the advantages of using LiDAR under different weather conditions. To explore the route in a GPS-free environment, \citet{malavazi2018lidar} introduced a method to extract lines from point cloud data. The approach is robust even when the plant condition is less than ideal. Furthermore, a LiDAR-based forest management strategy was proposed to monitor and assess various forest attributes, such as tree height growth, canopy volume, and cover estimate \cite{wulder2008role}. According to \cite{debangshi2022lidar}, combined with machine learning algorithms, data collected using LiDAR sensor can be applied to classify the species of plants. In addition, LiDAR is also used for forest fire prevention, soil erosion detection, yield prediction, etc.

\textbf{LiDAR Simulation.} LiDAR simulation \cite{manivasagam2020lidarsim, fang2020augmented} is a process that simulates the operation and performance of LiDAR systems that use laser beams to measure the distance and surface characteristics of targets. It is a very popular technique in many fields such as autonomous driving, robotics, and virtual reality. LiDARsim \cite{manivasagam2020lidarsim} uses real-world data, physical principles, and machine learning algorithms to simulate authentic LiDAR sensor data. \citet{fang2020augmented} presented an augmented LiDAR simulation system that automatically produces annotated point clouds for 3D obstacle perception in autonomous driving. And \citet{li2023pcgen} proposed a scalable simulation pipeline to transfer annotated point clouds across LiDAR embodiments, rendering synthetic data streams that emulate alternative densities and placements to generalize models.

\textbf{LiDAR-based Applications.} In addition to forestry, point cloud generated from LiDAR have huge potential in various tasks and domains~\cite{hasan2022lidar, wu2020deep, lu2021extending, kaartinen2022lidar, ghallabi2018lidar, he2023hierarchical, hu2023m}, especially in computer vision, augmented reality (AR) and virtual reality (VR). \citet{hasan2022lidar} indicates that LiDAR could be a powerful tool in computer vision tasks, such as object detection, person tracking and property estimation. \citet{ghallabi2018lidar} used multi-layer LiDAR data to detect lane markings, which were matched to a prior map using particle filtering to achieve improvements over standard GPS solutions. \citet{jacobsen2021real} proposed a novel worker safety monitoring system using LiDAR for precise real-time presence detection near hazards, demonstrably improving over GPS solutions when tested in a virtual environment. Furthermore, in \cite{gupta2014augmented}, the researchers introduced a mobile augmented reality platform which utilized LiDAR point cloud data to visually render real-world object dimensions on a phone.

\section{LiDAR-Forest Dataset}
\label{sec:dataset}
In this part, we present the methodology and specifics of the \textit{LiDAR-Forest} dataset. We've developed a procedure-based LiDAR Simulator to generate an authentic dataset. It's versatile and can be applied to various LiDAR sensors and configurations (e.g., rotation speed) with just a few key settings, such as scan pattern, beam quantity, and field of view (FOV). This procedure-based simulator follows the strict steps and accuracy of real scanning.

\subsection{Dataset Simulation}
We chose the Velodyne VLP-16 as a prototype for analog LiDAR, a widely chosen model in the current backpack LiDAR field, which has been similarly validated for stability and reliability by \cite{glennie2016calibration, lin2022comparative}. The overview of the framework is shown in Figure~\ref{fig:intro_fig}, which consists of 5 modules, the \emph{Data Generation Module}, the \emph{Rotation Module}, the \emph{Error Simulation Module}, the \emph{Human Movement Module}, and the \emph{Scanning Path Module}. We will detail each module function and the dataset format here. 

\emph{1) Data Generation Module}: The Data Generation Module involves simulating LiDAR beams for generating data in a virtual environment. Instead of directly replicating LiDAR beams, our approach is to create a prototype model and affix thin rectangles to it. These rectangles act as simulated beams and move in tandem with the base model, evenly distributed within a sector corresponding to the LiDAR's field of view (FOV). The length of each rectangle matches the maximum distance covered by a real LiDAR.

This strategy makes effective use of Unreal Engine 5's collision detection system. By employing this system, collision status between the rectangles and scene elements (such as trees or the ground) can be queried. When a collision is detected, the coordinates of the collision point are recorded, mirroring the process of a physical LiDAR.

The resulting data format encapsulates various key elements, including (x, y, z, leaf-wood labels, semantic labels, and instance labels), while (x, y, z) represents the coordinate of the collision point, leaf-wood labels refer to the type of object, which is used to indicate whether it is a leaf or a trunk, semantic labels refer to the category of the object, which is used to differentiate between different kinds of objects, e.g., ground, tree, stone, etc., and instance labels refer to the specific kind of individual of a semantic type, e.g., Tree A, Tree B, and Tree C. In the meantime, 2 types of data are generated: Relative and Absolute. The relative data type reflects LiDAR's perspective, detailing the coordinates relative to its position in the virtual space, and the absolute data type provides a broader context, offering real coordinates in the virtual space.
\\
\\
\emph{2) Rotation Module}: To adapt various scanning patterns, we introduced a versatile rotation configuration employing three key parameters within our research. These parameters encompass azimuth resolution, vertical angles, and spin rate.

\begin{figure}[htb]
\centering
\includegraphics[width=0.85\linewidth, height=6.5cm]{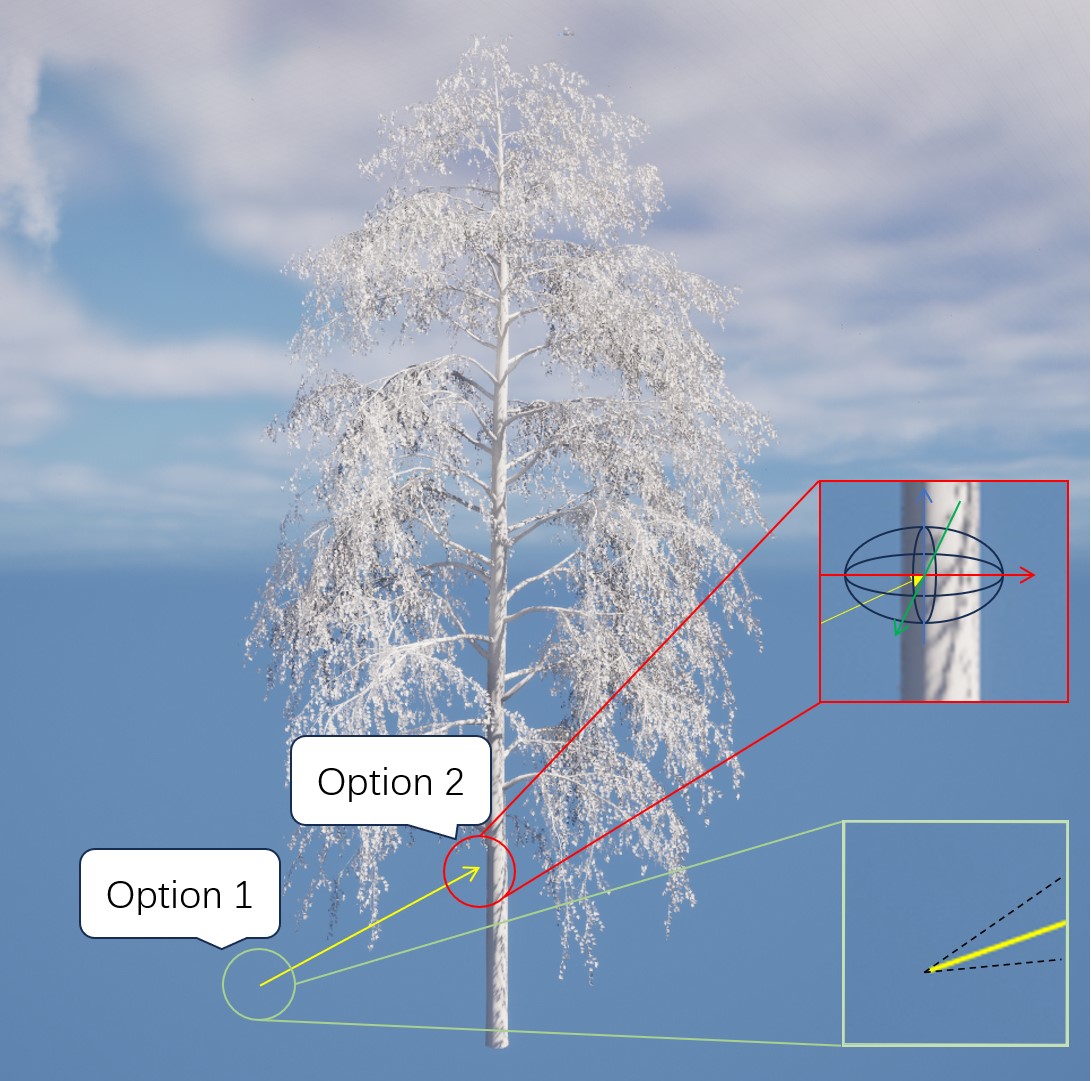}
\vspace{-2mm}
\caption{{Specific embodiment of the two error simulation methods: \textit{Option 1} corresponds to the scheme that modifies the angle, and \textit{Option 2} corresponds to the scheme modifying the coordinates directly.}
\label{fig:error}}
\vspace{-3mm}
\end{figure}

Starting with azimuth resolution, this parameter defines the level of detail in the horizontal scanning dimension. By adjusting the azimuth resolution, we can customize the precision of scans based on the specific demands of the environment or application. The second parameter, vertical angles, plays a crucial role in determining the extent of coverage in the vertical dimension. This flexibility allows us to tailor the scanning process to different spatial requirements. Whether it's capturing data from specific heights or obtaining a comprehensive vertical profile, adjusting vertical angles enables us to align the scanning methodology with the objectives of the given task. Lastly, the spin rate parameter governs the speed at which the scanning device rotates. This dynamic control over the spin rate is essential for optimizing scanning efficiency. Depending on the application, a faster or slower spin rate can be employed to strike a balance between swift data acquisition and the precision required for accurate analysis.
\\
\\
\emph{3) Error Simulation Module}: The Error Simulation Module (see Fig.~\ref{fig:error}) is crucial for emulating the Velodyne VLP-16 LiDAR sensor. We refer to the VLP-16 User Manual for insights into beam divergence. However, data phasing challenges prompt the need for a segmentation function. This function, proposed within our module, accurately simulates beam divergence despite data phasing.

Addressing the utilization of error queried from the segmentation function involves exploring 2 proposed methods. One approach is to calculate the error at the conclusion of the process, incorporating it by adding an offset to the collision point's coordinates. It assumes the collision point must inside a ellipsoidal space whose central point is the collision point, it's semi-long axis' length is the horizontal divergence and short axis's length is the vertical divergence. Under this hypothesis, value of the offset should be a random value pair (x, y) with constraint that in the ellipsoidal space.

Alternatively, an initial adjustment to the beam's angle based on the error is suggested, leading to the determination of coordinates with the incorporated error. The angle has 2 parts, vertical angle and horizontal angle. In order to calculate this angle, we need to get the distance between the starting position (coordinate of the prototype model) and the collision point, through this distance value we can know the maximum value of divergence at the current stage from the segmentation function, divide the distance by divergence to get the tangent value, and then use the inverse trigonometric function to get the required angle.
\\
\\
\emph{4) Human Movement Module}: Mobile Mapping Systems (e.g., backpack or hand-held LiDAR systems) have been used in the real forest scene for faster data acquisition~\cite{lin2022comparative}. Our study simulates this scenario where a person is walking through a forest with a LiDAR device on his back, so we need to take into account the vertical and horizontal swaying that occurs during a person's walk, even if it is only slight, when simulating. To achieve a more realistic dataset in the context of this scene, we suggest a human movement simulation into the system. 

For simplicity, we divide human movement into two primary components: vertical and horizontal movement. The vertical movement, resembling an up-and-down motion, and the horizontal movement, capturing side-to-side swaying, can both be effectively represented using sine functions. This simplification allows us to inject a level of realism into our simulations, mirroring the intricacies of human motion as it influences LiDAR data acquisition in a forest environment.
\\
\\
\emph{5) Scanning Path Module}: This module is designed with two primary functionalities: assigning the scanning route and adjusting the running time (total frames) in accordance with the spinning rate and azimuth resolution.

In the process of assigning the scanning route, we leverage the intuitive capabilities of the spline component in UE5. This component facilitates a user-friendly approach, allowing for the seamless adjustment of the scanning path through a drag-and-drop interface. %The spline component proves to be a versatile tool for creating and modifying scanning routes, providing flexibility in adapting to specific simulation requirements. % 
On the other hand, the adjustment of running time poses a unique challenge due to the frames per second (FPS) limitations inherent in computer systems. Unlike real LiDAR devices that can achieve Rotation per Minute (RPM) speeds as high as 1200 rpm, computer-based FPS data acquisition is constrained. Assuming a fixed 30 frames per second, and considering the Velodyne VLP-16's User Manual specification of 8 data points collected per circle per beam, totaling 160 data points per second per beam, the required adjustment factor is derived from the ratio 160 / 30. This adjustment ensures that our simulated scanning time aligns proportionally with the operational characteristics of the Velodyne VLP-16, offering a realistic representation within the computational constraints of the simulation environment.

\subsection{Metrics and Evaluations}
\textbf{Metrics.} To evaluate the LiDAR simulation (\eg LiDAR types, placement, and scene complexity), we plan to follow~\cite{cai2023analyzing} to use two point cloud distribution metrics: Infrastructure Density (InfraD) and Infrastructure Normalized Uniformity Coefficient (InfraNUC). 

InfraD describes the density of the point cloud within the specific region of interest (called InfraLOB) to perceive as:
\begin{equation}
InfraD = \frac{N}{S}
\end{equation}

\noindent where $N$ is the number of points inside the InfraLoB regions and $S$ is the area of InfraLoB in the Unreal Engine.

As a variant of Normalized Uniformity Coefficient (NUC) \cite{li2019pu}, InfraNUC is a measure of the overall uniformity of the point set across all target objects. It is calculated as the standard deviation of the points in the randomly selected disk regions as:
\begin{equation}
\begin{split}
\textit{InfraNUC} &= \sqrt{\frac{1}{D} \sum_{i=1}^D\left(\frac{n_i}{N \cdot p}-\frac{1}{D} \sum_{i=1}^D \frac{n_i}{N \cdot p}\right)}
\end{split}
\end{equation}

\noindent where $n_{i}$ is the number of points inside the disk regions, $N$ is the number of points within the InfraLOB region, $D$ is the total number of disks, and $p$ is the ratio of disk area to InfraLOB area.

Beyond the metrics for point cloud itself, there are more metrics we will consider to include in relevant applications from the dataset, such as Mean average precision (mAP) for calculating each detected objects overlapping over the specified categories for the task of 3D tree detection, and Chamfer distance \cite{li2019pu} for measuring the similarity of the two sets of point clouds $P_{est} \in R^{N\times3}$ and the ground truth one $P_{gt} \in R^{N\times3}$ in point cloud reconstruction and interpolation.

\begin{figure*}[htb]
\centering
\includegraphics[width=0.96\linewidth, height=6.5cm]{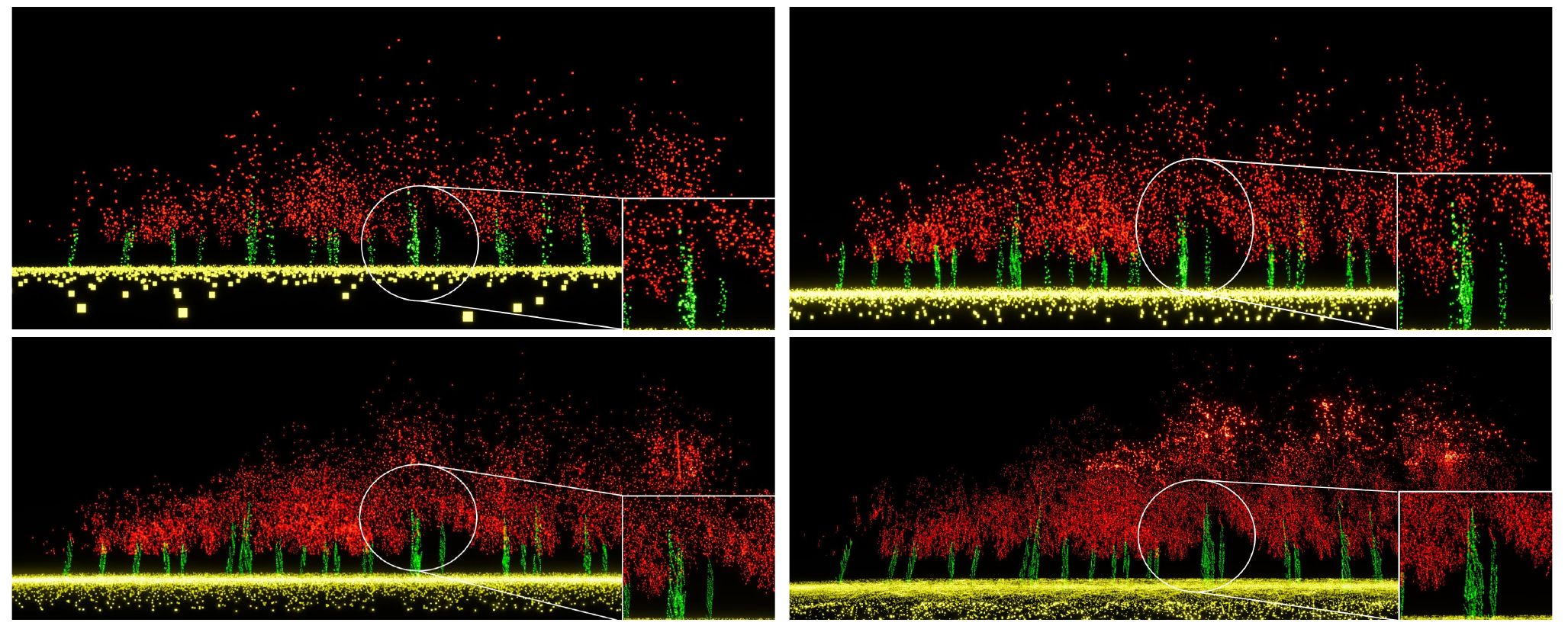}
\vspace{-2mm}
\caption{Illustration of the simulation results from different types of LiDAR sensors in our \textit{LiDAR-Forest} dataset. Top left: simulation result from 8-bit LiDAR; Top right: simulation result from 16-bit LiDAR; Bottom left: simulation from 64-bit LiDAR; Bottom right: simulation from 256-bit LiDAR. All the given point clouds are generated from the same scenario and the same viewpoint using different sensor patterns.
\label{fig:result}}
\vspace{-3mm}
\end{figure*}

\textbf{Evaluations.}
In Figure \ref{fig:result}, we illustrate the qualitative result from different types of LiDAR sensor data (VLD-8 bit, VLD-16 bit, VLD-64 and VLD-256 bit), where all the background is from the same scenario and the foreground objects are placed at the same location. It can be observed that as the bit increases, the point becomes denser and more detailed. For a better visualization, we crop a small part and zoom it in to provide the detailed point structure of the object.

\section{Potential Applications}
\label{sec:usage}

The \textit{LiDAR-Forest} dataset provides non-error labels, which allows us to train data-intensive algorithms for various forestry tasks and evaluate them using the high-quality ground truth~\cite{xu2023spatial,lin2022comparative,lu2023label}.

\textbf{Tree Species Identification.} In field scanning, species information needs a forest professional to recognize one by one, which is not beneficial for fast ground truth collection. Tree models in our simulation process can be customized for any species, so the species information is directly known without identification. This advantage is helpful in species ID dataset creation and algorithm development.

\textbf{Stem Mapping and Measuring.} Forest inventory is a basic but important task in forestry practice, and it includes stem mapping (counting and localization) and attributes measuring (DBH and tree height). Due to instance-level labels are provided in our dataset, we can easily conduct algorithm development and evaluation for individual tree segmentation that can be used for stem mapping. With a simulated dataset, one can generate forest point clouds with each tree's location, DBH, and height. Furthermore, key data attributes such as point density, noise level, and occlusion can be controlled as well, which helps evaluate how different scanning conditions and point cloud quality affect the accuracy of tree diameter estimation.

\textbf{Volume and Biomass Estimation.} Derivation of tree volume and biomass from point cloud requires wood points, but it is hard to separate wood points from the tree point cloud as small branch points and foliage points are often stuck together. Thereby, obtaining accurate leaf-wood labels from real point clouds is almost impossible. Thanks to our simulator, we can generate no-error leaf-wood labels as wood and leaf are known. Accurate leaf-wood labels help build more reliable algorithms for tree volume estimation. With species information from our dataset, biomass can also be calculated.

\textbf{Digital Twinning and Simulation for Virtual Experiences.} Other than ecological attribute measurement, our simulated dataset can also be used for forest scene reconstruction of digital twins. The reconstructed forest provides valuable insights into ecological processes, forest dynamics, and environmental changes. The detailed reconstructions can be used for VR and AR applications, offering immersive experiences for education, training, or entertainment.

\vspace{-1mm}
\section{Future works}
\label{sec:future}

For future work, we plan to use the simulated dataset in multimodal sensor fusion. This will combine the simulated point cloud with additional synthesized RGB imagery, depth maps, and other modalities for multi-sensor perception research. 

\vspace{-1mm}
\section{Conclusion}
\label{sec:conclusion}
This paper introduces the \textit{LiDAR-Forest Dataset}, which addresses the lack of good dataset for LiDAR point cloud simulation in wild forests. To establish the simulation platform, we present asset-scene-simulation stages for setting up a complex forest simulator. To replicate real-world LiDAR sensor effects, we propose five modules, including sensor error and human movement simulations. We will
release the dataset to the public, hoping to promote related research in 3D point cloud technology for forestry, engineering, and education communities. In the future, our dataset will support additional types of sensors. 
LiDAR sensors and more learning-based perception tasks like point cloud segmentation, reconstruction, and interpolation.

\vspace{-1mm}
\section{Acknowledgments}
This ongoing work is supported by the U.S. Department of Agriculture (USDA) under grant No. 20236801238992.

\clearpage

\bibliographystyle{abbrv-doi}

\bibliography{template}
\end{document}